\documentclass[lettersize,journal]{IEEEtran}
\usepackage{amsmath,amsfonts}
\usepackage{algorithmic}
\usepackage{algorithm}
\usepackage{array}
\usepackage[caption=false,font=scriptsize,labelfont=sf,textfont=sf]{subfig}
\usepackage{textcomp}
\usepackage{stfloats}
\usepackage{url}
\usepackage{verbatim}
\usepackage{graphicx}
\usepackage{cite}
\usepackage{multirow}
\usepackage{arydshln}
\usepackage[export]{adjustbox}

\begin{document}

\title{Prompt Guided Transformer for Multi-Task Dense Prediction}

\author{Yuxiang Lu,~\IEEEmembership{Student Member, IEEE}, Shalayiding Sirejiding,~\IEEEmembership{Student Member, IEEE},

Yue Ding,~\IEEEmembership{Member, IEEE}, Chunlin Wang, and Hongtao Lu,~\IEEEmembership{Member, IEEE}
\thanks{Yuxiang Lu, Shalayiding Sirejiding, Yue Ding, and Hongtao Lu are with the Department of Computer Science and Engineering, Shanghai Jiao Tong University, Shanghai 200240, China (email: luyuxiang\_2018@sjtu.edu.cn; salaydin@sjtu.edu.cn; dingyue@sjtu.edu.cn; htlu@sjtu.edu.cn) }
\thanks{Chunlin Wang is with the School of Information Science and Technology, Chuxiong Normal University, Chuxiong 675099, China (email: wcl@cxtc.edu.cn)}
}

\markboth{Journal of \LaTeX\ Class Files,~Vol.~14, No.~8, August~2021}%
{Y. Lu, S. Sirejiding, \MakeLowercase{\textit{et al.}}:Prompt Guided Transformer for Multi-Task Dense Prediction}

\maketitle

\begin{abstract}
    Task-conditional architecture offers advantage in parameter efficiency but falls short in performance compared to state-of-the-art multi-decoder methods.
    How to trade off performance and model parameters is an important and difficult problem.
    In this paper, we introduce a simple and lightweight task-conditional model called Prompt Guided Transformer (PGT) to optimize this challenge. Our approach designs a Prompt-conditioned Transformer block, which incorporates task-specific prompts in the self-attention mechanism to achieve global dependency modeling and parameter-efficient feature adaptation across multiple tasks. This block is integrated into both the shared encoder and decoder, enhancing the capture of intra- and inter-task features.
    Moreover, we design a lightweight decoder to further reduce parameter usage, which accounts for only 2.7\% of the total model parameters. 
    Extensive experiments on two multi-task dense prediction benchmarks, PASCAL-Context and NYUD-v2, demonstrate that our approach achieves state-of-the-art results among task-conditional methods while using fewer parameters, and maintains a significant balance between performance and parameter size. 
\end{abstract}


\section{Introduction}
The human brain and its visual system are remarkable in their ability to perform multiple tasks.
Similarly, Multi-Task Learning (MTL) emerges as a popular topic in machine learning and computer vision, with the goal of training a single model to tackle multiple related tasks simultaneously.
MTL capitalizes on the idea of learning shared representations across tasks, capturing common and complementary information, which brings more accurate performance \cite{survey}.
Moreover, the parameter sharing mechanism makes learning a multi-task model more efficient than training single-task models separately.

\begin{figure}
    \centering
    \subfloat[Multi-decoder]{\includegraphics[width=0.3\linewidth]{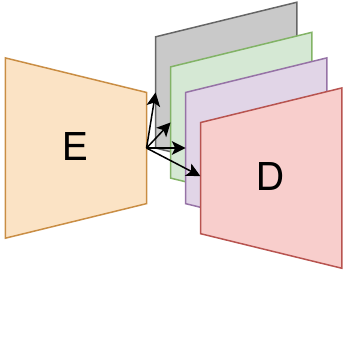}\label{fig:arch1-md}}
    \hspace{1mm}
    \subfloat[Task-conditional]{\includegraphics[width=0.3\linewidth]{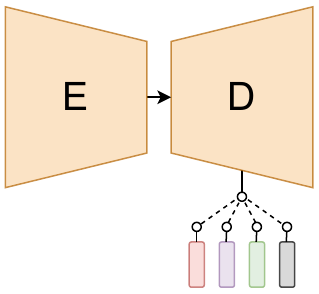}\label{fig:arch1-tc}}
    \hspace{1mm}
    \subfloat[PGT (Ours)]{\includegraphics[width=0.3\linewidth]{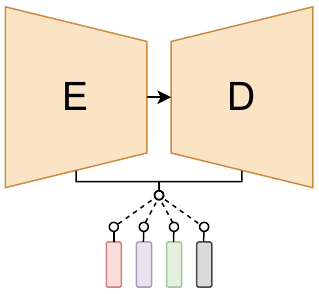}\label{fig:arch1-pgt}}
    \caption{Different architectures of MTL methods. (a) Conventional multi-decoder approach uses a shared encoder and multiple task-specific decoders. (b) Task-conditional method uses a single encoder-decoder architecture with task-specific modules in decoder. (c) Our proposed PGT adopts task-conditioning strategy on both the encoder and decoder.}
    \label{fig:arch1}
\end{figure}

Dense prediction tasks, \textit{e.g.} semantic segmentation, edge detection, depth estimation, are fundamental tasks in the field of computer vision. Recently, addressing various dense prediction tasks via an MTL framework has been a rising topic.
Encoder-focused methods \cite{cross-stitch, nddr-cnn, mtan} and decoder-focused methods \cite{pad-net, mti-net, atrc, invpt} are two prominent technical routes, both of which intend to learn common representations and cross-task relationships. 
Generally, as shown in Fig. \ref{fig:arch1-md}, these methods learn individual decoders for each task and design the decoders with a complex network structure, thus suffering from a high number of parameters. Such drawback will be significantly amplified when applying separate decoders to handle more tasks.

\begin{figure}[t]
    \centering
    \vspace{2mm}
    \includegraphics[width=\linewidth]{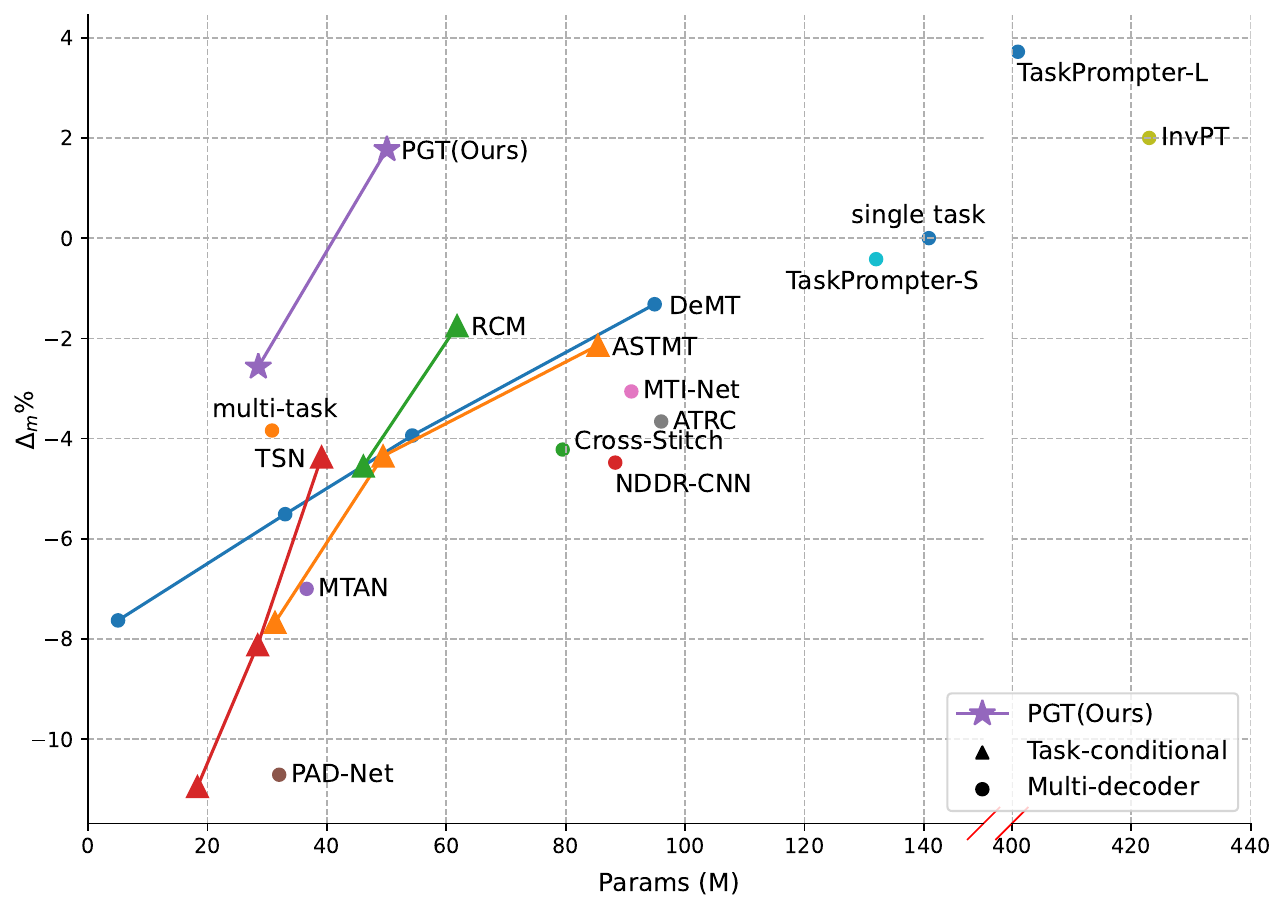}
    \caption{Multi-task performance $\Delta_m$ (w.r.t. single task baseline) \textit{v.s.} number of parameters on PASCAL-Context dataset. Our PGT achieves a remarkable balance compared with state-of-the-art task-conditional and multi-decoder methods.}
    \label{fig:comp}
\end{figure}

An alternative MTL architecture is based on the task-conditional paradigm \cite{astmt, rcm, tsn, composite}. Different from conventional methods that generate results for several tasks at one time, task-conditional methods perform separate forward passes within the model for each task. This paradigm offers the advantage of parameter efficiency and a flexible architecture, which makes it practical in many real-world scenarios \cite{rcm}. Among these methods, TSN \cite{tsn} makes great progress in solving the issue of parameter expansion by sharing the decoder in the network, as depicted in Fig. \ref{fig:arch1-tc}. 

However, there exists a performance gap between task-conditional methods and encoder- or decoder-focused methods. State-of-the-art multi-decoder methods \cite{vpt1, invpt, mqt, demt, mtformer, taskprompter} largely improve performance by leveraging powerful Vision Transformer (ViT) models, which are proved more appropriate for MTL than convolutional neural networks (CNNs). Nevertheless, it is difficult to apply transformer-based models to task-conditional architecture while maintaining its advantage in parameter efficiency, since the attention mechanism considerably increases the number of parameters. Additionally, designing an effective conditioning strategy poses a challenge as it performs task interaction implicitly.
On the other hand, existing task-conditional approaches are limited in jointly learning shared, task-specific and cross-task information on both the encoder and decoder in a lightweight manner, thereby restricting their representation learning capability for multiple tasks.

To address the aforementioned limitations, we propose a novel method called Prompt Guided Transformer (PGT) for multi-task dense prediction. Our goal is to improve multi-task performance while mitigating the excessive utilization of parameters in existing works. As shown in Fig. \ref{fig:comp}, we compare the overall performance and number of parameters between representative MTL methods, including both multi-decoder and task-conditional architectures. 
Our method achieves a remarkable balance between performance and parameter efficiency. 
Notably, our PGT yields comparable results to state-of-the-art transformer-based models, namely InvPT \cite{invpt} and TaskPrompter \cite{taskprompter}, while using considerably fewer parameters (note the gap on the x-axis).
Moreover, our model size is less susceptible to the number of tasks compared to existing methods (see Fig. \ref{fig:param}).

Inspired by the great success of prompting in transfer learning \cite{prompt_survey, pt1, vpt_prompt}, we introduce a conditioning strategy based on prompts, which empowers our transformer-based model to effectively learn both intra- and inter-task knowledge with the combination of task-specific prompts and shared attention weights.
We follow the task-conditional multi-tasking to ensure efficiency, where a set of task prompts are activated during the forward process to adapt the features for each specific task.
We argue that most of existing MTL methods employ a shared encoder for multiple tasks, overlooking the crucial fact that distinct tasks necessitate diverse features extracted from the encoder. Consequently, as illustrated in Figure \ref{fig:arch1-pgt}, we apply our conditioning strategy to both the encoder and decoder, thereby enhancing the model's ability to acquire task-specific features.
Specifically, we design a Prompt-conditioned Transformer (PcT) block to capture long-range relationships within feature maps. Under the guidance of task prompts, the PcT block can learn task-specific representations explicitly, model common information shared among tasks, and capture cross-task interactions implicitly through shared parameters.
Moreover, we integrate the PcT block into a task-conditional framework and design a lightweight PGT decoder. The task-specific parameters occupy only 1.25\% of the overall model parameters, resulting in an efficient multi-task learning approach.

In summary, the main contributions of this paper are as follows:
\begin{itemize}
    \item We propose Prompt Guided Transformer (PGT) with a conditioning strategy based on task-specific prompts. PGT leverages the advantage of transformer model in capturing long-range interactions and a lightweight task-conditional framework to alleviate the dilemma between performance and parameters.
    \item We design a Prompt-conditioned Transformer (PcT) block as an effective conditioning strategy for both the encoder and decoder, which facilitates the joint learning of task-specific, common, and cross-task information. Additionally, it enables efficient feature adaptation across tasks while minimizing the number of parameters required.
    \item Experimental results on two widely-used benchmark datasets, PASCAL-Context and NYUD-v2, show that our method remarkably surpasses previous task-conditional methods by nearly \textbf{2\%} and \textbf{10\%} respectively, establishing a superior balance between accuracy and network size.
\end{itemize}

\begin{figure*}[t]
    \centering
    \includegraphics[width=\textwidth]{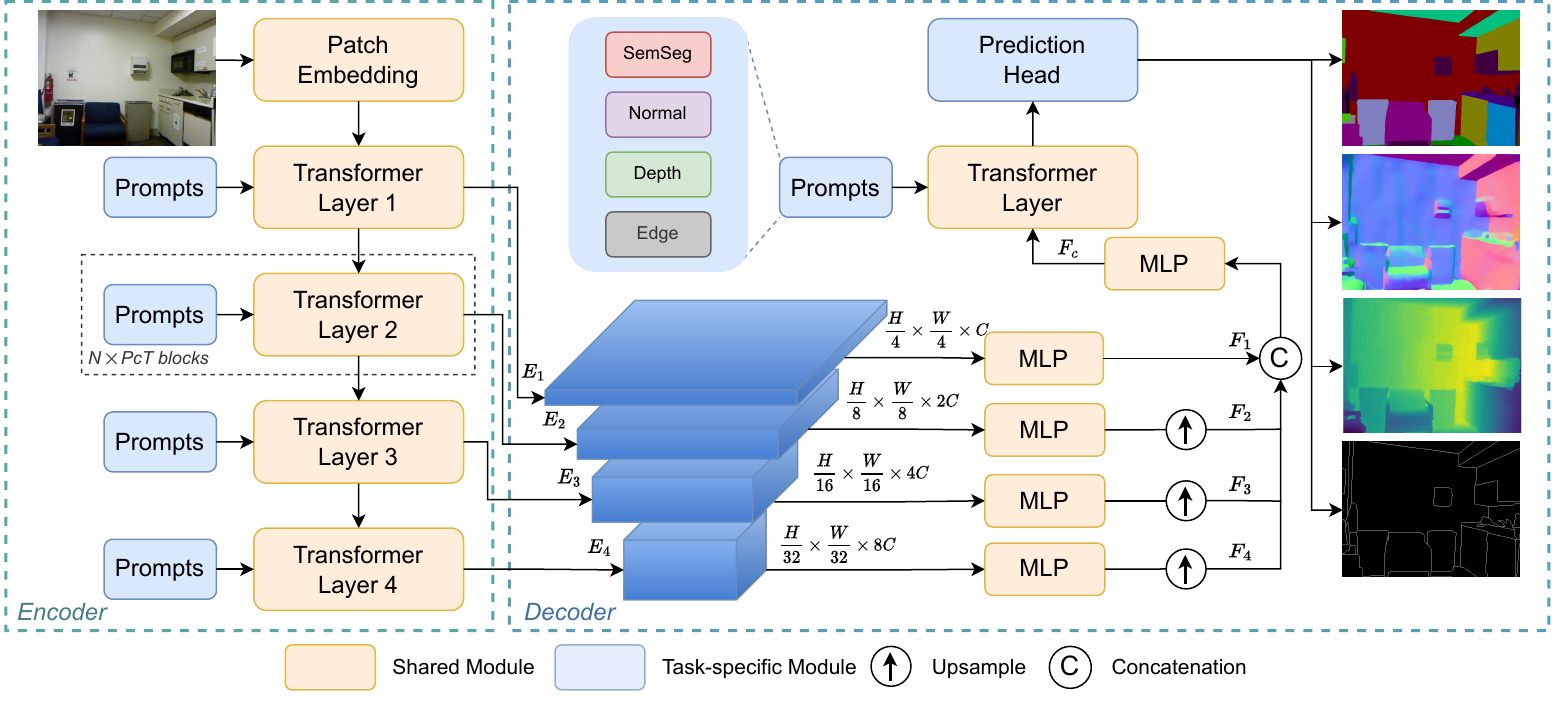}
    \caption{Overview of the proposed Prompt Guided Transformer (PGT). PGT is based on task-conditional architecture, which divides the network into shared and task-specific modules. For an input image, PGT encoder extracts multi-scale features using a four-layer transformer backbone. These feature maps are then processed by MLP modules and upsampling operations before concatenation. A transformer layer is used to decode the fused feature and task-specific prediction heads are used to generate final results. The Prompt-conditioned Transformer (PcT) block is used to build all transformer layers, where a set of corresponding prompts are activated during the forward passes for each task.}
    \label{fig:arch}
\end{figure*}

\section{Related Works}

\paragraph{Multi-Task Learning for dense prediction} 
Multi-Task Learning (MTL) aims to improve overall performance while reducing parameters and speeding up training or inference \cite{mtl1,mtl2,mtl3}. 
Existing MTL methods focus on learning shared and specific representations through approaches such as parameter sharing \cite{taps, dmtrl, sfg, tmm_mt1}, task interaction \cite{mrn, mtformer, jtrl, psd, vpt1, tmm_mt2}, and prediction distillation \cite{pad-net, pap-net}. 

MTL architectures can be roughly categorized into encoder-focused and decoder-focused methods based on where information is exchanged or shared between tasks \cite{survey}. Encoder-focused models use shared encoders to extract general features, with mechanisms such as feature fusion \cite{cross-stitch, nddr-cnn, sluice}, attention \cite{mtan}, dynamic branching \cite{fafs, branched, bmtas, ltb}. Decoder-focused models first make initial predictions for each task and then refine the results by extracting inter-task correlation with information distillation modules \cite{pad-net, pap-net, mti-net, atrc, invpt}. Though these methods achieve impressive accuracy, they suffer from the large network scale.

\paragraph{Task-conditional architecture}
Recently, another MTL architecture is the task-conditional paradigm, which introduces task-specific modules in the main stem of the network. During the separate forwarding of each task, corresponding modules are activated to achieve feature adaptation from the shared modules. ASTMT \cite{astmt} and RCM \cite{rcm} perform task-conditioning on encoders with attention modules and convolution reparameterization, respectively. TSN \cite{tsn} proposes a single-encoder-single-decoder architecture by learning task-conditional parameters with a task embedding network. \textit{Popovic} \textit{et al.} \cite{composite} extend the architecture of TSN to perform multiple, spatially distributed tasks.
In this paper, we propose PGT to enhance the conditioning strategy with prompting and promote its advantage in parameter efficiency.

\paragraph{Vision Transformer}
Transformer and the self-attention mechanism are initially introduced in natural language processing (NLP) and have achieved significant success in other fields such as computer vision \cite{transformer}. Vision Transformer (ViT) leverages self-attention modules to capture long-range dependencies among image patches \cite{vit}. ViT-based models demonstrate remarkable performance on various vision tasks, such as image classification \cite{vit, swin, tmm_tr1, tmm_tr2}, object detection \cite{detr, deformabledetr}, dense prediction \cite{ipt, dpt, pvt, segmentor}, and video understanding \cite{videotr, videotr2}, surpassing previous CNN-based models. Recently, transformer models have been extended to multi-task dense prediction to learn better features and model more accurate task relationships \cite{vpt1, invpt, mqt, demt, mtformer, taskprompter, tet}. Our PGT differs from these methods in that we design a transformer-based model with a lightweight task-conditional framework, providing a promising solution to address the issue of high parameter counts.

\paragraph{Prompting}
\textit{Prompting} is a technique that inserts language instructions into the input space to help a transformer-based language model understand a task and adapt pre-trained models to downstream tasks by only fine-tuning a small portion of parameters \cite{prompt_survey}. Some recent works improve prompts by treating them as learnable task-specific vectors in the input token sequence and updating them through back-propagation. These works are collectively referred to as Prompt Tuning methods \cite{pt1, pt2, pt3, hyperprompt}. Compared with full fine-tuning that updates all the parameters, prompt tuning is more efficient but also achieves comparable performance.
Prompting is also applied to vision tasks for transfer learning \cite{vp, vpt_prompt, evp} or continual learning \cite{pt_cont1, pt_cont2}. For example, VP \cite{vp} uses image perturbations as prompts in the pixel space, while VPT \cite{vpt_prompt} introduces a small number of learnable tokens into the ViT model. Unlike transfer learning which intends to fill the gap between various data domains, we use prompting to bridge the gap among multiple task domains, which is a parameter-efficient method to perform feature adaptation. 
More related to our work, TaskPrompter \cite{taskprompter} uses task prompt to learn task-specific representations in multi-task dense prediction. It differs from our proposed method in that TaskPrompter simultaneously learns spatial and channel task prompts in one forward pass, whereas our PGT adopts a task-conditional setting and utilizes task prompts to guide feature adaptation.

\section{Method}
In this section, we first introduce the problem formulation of multi-task learning and task-conditional setting. Afterward, we give an overview of the architecture of our proposed PGT and provide detailed illustrations of the Prompt-conditioned Transformer block, the PGT encoder, and the lightweight PGT decoder.

\subsection{Problem Formulation}
Given an RGB image $\mathbf{X}\in \mathcal{X}=[0, 255]^{H \times W \times 3}$ and $N$ tasks $T=\{t_1, t_2, \dots, t_N\}$, it is assumed that each task has an expected result $\mathbf{Y}_t\in \mathcal{Y}_t$. Concerning dense prediction tasks, the expected results should have the same size in height and width as the input image but may have different numbers of channels. For example, the shape of the surface normal estimation result is $H \times W \times 3$, while for edge detection, it is $H\times W \times 1$.

We define $f:\mathcal{X}\times T\times \Theta \rightarrow \bigcup_{t\in T} \mathcal{Y}_t$ as the multi-task learning function, where $\Theta$ is the model parameters. Hence, for the input $\mathbf{X}$, we have 
\begin{equation}
    f(\mathbf{X}, T, \Theta)=\bigcup_{t\in T} \mathbf{Y}_t.
\end{equation}

Considering the parameter sharing mechanism in MTL, we can divide the model parameters into two parts, 
\begin{equation}
    \Theta = \Theta_s + (\Theta_1 + \cdots + \Theta_N), \label{eq2}
\end{equation}
where $\Theta_s$ represents the parameters shared among all tasks and $\Theta_t$ is the task-specific parameters for task $t$ (equivalent number for each task). 

In task-conditional setting, each task is performed by a separate forward pass, which is alternatively denoted by
\begin{equation}
    f(\mathbf{X}, t, \Theta_s + \Theta_t)=\mathbf{Y}_t, \forall t\in T,
\end{equation}
where $\Theta_s + \Theta_t$ means that the task-specific modules corresponding to task $t$ are activated and combined with the shared stem.

Following previous work \cite{tsn}, we can assume the following condition to simplify the analysis,
\begin{equation}
    |\Theta_s| + |\Theta_t| \approx c, \forall t \in T. \label{eq4}
\end{equation}
where $|\Theta|$ represents the number of parameters.
Then from Eq. \ref{eq2} and Eq. \ref{eq4}, we get
\begin{equation}
    |\Theta|=|\Theta_s|+N|\Theta_t| = c+(N-1)|\Theta_t| \label{eq5}
\end{equation}

It is apparent from Eq. \ref{eq5} that the total number of parameters is directly proportional to the number of tasks. As the number of tasks increases, the total number of task-specific parameters also increases, occupying a substantial proportion of the overall model parameters. This leads to models with multiple task-specific decoders becoming relatively large.
Meanwhile, Eq. \ref{eq5} highlights that when the number of tasks remains constant, the number of parameters becomes linear to the number of shared parameters, encouraging us to use fewer task-specific parameters to obtain a lightweight model.

Task-conditional methods typically allocate only a tiny fraction of parameters to task-specific modules, with most of the network shared among the tasks. This results in a small $|\Theta_t|$ and a hefty $|\Theta_s|$, making it advantageous in reducing the number of parameters. In this paper, we extend the task-conditional method to transformer-based models for performance improvement and incorporate the concept of prompting to further reduce the number of task-specific parameters and the total number of parameters.

\subsection{Prompt Guided Transformer}

\paragraph{Overview}
As depicted in Fig. \ref{fig:arch}, our proposed model follows an encoder-decoder architecture. The input to the model is an RGB image and a specified task type, and the encoder leverages a transformer backbone to extract multi-scale features at four hierarchical levels.
Our model employs task-specific prompts and integrates them into the transformer layers. The decoder fuses the features adaptively with dimensionality reduction and uses a transformer layer to decode the fused features, and similarly, we integrate task-specific prompts into the decoder. Finally, the task-specific prediction heads produce results for each task. 
We call the transformer block with prompts as Prompt-conditioned Transformer (PcT) block, which adapts general features to each task domain.

\begin{figure}[t]
    \centering
    \includegraphics[width=\columnwidth]{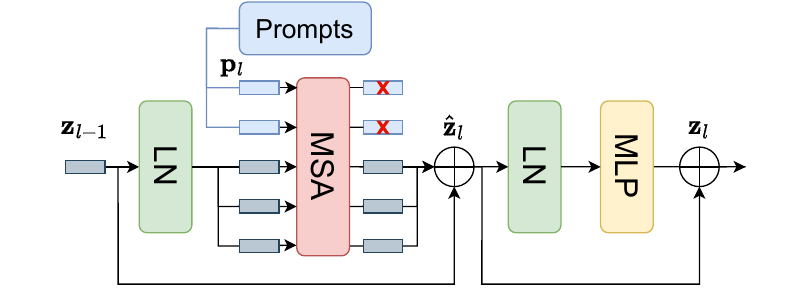}
    \caption{Prompt-conditioned Transformer (PcT) block. The task-specific prompts are inserted into the token sequence before the Multi-head Self Attention (MSA) module and then separated after MSA.}
    \label{fig:block}
\end{figure}

\paragraph{Prompt-conditioned Transformer block}
The Prompt-conditioned Transformer block is a critical component of our proposed model, enabling feature representations to be conditioned on the encoder and decoder to satisfy the requirements of a specified task. Fig. \ref{fig:block} depicts the structure of the PcT block, following ViT models \cite{vit, swin}. The input feature tokens are first processed by a LayerNorm (LN) layer and then concatenated with prompts before inputing to a Multi-head Self Attention (MSA) module, which enables long-range dependency modeling to capture intra-task information. The MSA module is followed by another LN layer and a two-layer Multi-Layer Perceptron (MLP) module. Also, residual connections are performed after the MSA and MLP modules.

To facilitate the implementation of task-specific prompts, we draw inspiration from VPT \cite{vpt_prompt} and embed them into the input sequence of the MSA module. For each task, the prompts $\mathbf{p}_l\in \mathbb{R}^{N_p\times d}$ constitute a set of $N_p$ tokens of dimension $d$, equivalent to a feature token's dimension. The task-specific representations are learned by activating different task prompts, while the universal information and cross-task correlation are implicitly modeled by the parameters shared among all tasks.
Moreover, to reduce computational costs, the prompts are separated after the self-attention operation, with only the feature tokens undergoing subsequent processing in the MLP module.

 The whole process of our Prompt-conditioned Transformer block is formulated as
\begin{align}
    [\mathbf{p}_l'&, \mathbf{z}_l']=\text{MSA}(\text{Concat}(\mathbf{p}_l, \text{LN}(\mathbf{z}_{l-1}))),\\
    \hat{\mathbf{z}}_l&=\mathbf{z}_l'+\mathbf{z}_{l-1}, \\
    \mathbf{z}_l&=\text{MLP}(\text{LN}(\hat{\mathbf{z}}_l))+\hat{\mathbf{z}}_l.
\end{align}
where $\mathbf{p}_l'\in \mathbb{R}^{N_p\times d}$ and $\mathbf{z}_l'\in \mathbb{R}^{L\times d}$ represent the output prompts and features of the MSA module, respectively, while $\mathbf{z}_l\in \mathbb{R}^{L\times d}$ denotes the output tokens of the block $l$.

As depicted in Fig. \ref{fig:arch}, the PcT block is utilized as the basic module to build transformer layers in the encoder and decoder, with unique prompts inserted into every block for each task. As multi-task dense prediction involves both low-level tasks and high-level tasks, applying distinct prompts at varying levels is essential. The multi-level prompts can assist the network in comprehending the current task and allow for improved task-specific feature extraction.

\paragraph{PGT encoder}
In order to improve feature extraction and promote long-range dependency modeling, we adopt well-designed Vision Transformer backbones in our encoder. First, the input image is split into non-overlapping patches and projected into tokens of dimension $C$ in the patch embedding layer. Then hierarchical presentations $E_1$ to $E_4$ are generated from layer 1 to layer 4 with self-attention computation. The number of tokens is reduced by $1/4$, and the dimension is doubled with the patch merging operation after the first three layers. Therefore, the output resolutions of the encoder range from $\frac{H}{4}\times \frac{W}{4}\times C$ to $\frac{H}{32}\times \frac{W}{32} \times 8C$, which is suitable for vision tasks of different levels, and also consistent with previous CNN-based backbones for dense prediction such as ResNet \cite{resnet} and HRNet \cite{hrnet}.

\paragraph{Lightweight PGT decoder}
Multi-task models that adopt the transformer-based encoder typically have more parameters than those that employ the CNN-based encoder. Intuitively, we should pay attention to parameter utilization in the decoder. Unfortunately, previous methods overlook it and usually adopt complex network structures for the decoder. In this work, we simplify the decoder because our method has a strong capability of task domain adaptation with the help of the PcT blocks. 

To facilitate a lightweight decoder, it is critical to consider the dimension of the features. Therefore, we first apply feature fusion to adaptively combine features $E_1$ to $E_4$, see Fig. \ref{fig:arch}, and conduct dimensionality reduction. Specifically, we pass the features through MLP modules to project them onto a lower dimension $C$. The output features from layer 2 to layer 4 are then upsampled to match the spatial resolution of layer 1, which is $\frac{H}{4}\times \frac{W}{4}$. Afterward, the four feature maps $F_1$ to $F_4$ are concatenated to form a fused feature map and processed by another MLP module to reduce the dimension to $C$ once again. We can formulate the feature fusion by the following equations,
\begin{align}
    F_i&=\left\{\begin{array}{ll}
        \text{MLP}(E_i), &  i=1\\
        \text{Upsample}(\text{MLP}(E_i)), & i=2,3,4
    \end{array}
    \right. \\
    F_c&=\text{MLP}(\text{Concat}(F_1, F_2, F_3, F_4))
\end{align}
where $F_c$ is the fused feature map of shape $\frac{H}{4}\times \frac{W}{4}\times C$.

After the feature fusion and dimensionality reduction, the fused feature is processed through a single transformer layer to decode task-specific features while preserving its small shape. Finally, it is fed into task-specific prediction heads to output results for each task. One prediction head consists of two upsampling layers and an $1\times 1$ convolution layer for channel projection. 

\section{Experiments}
In this section, we present extensive experiments to evaluate our proposed PGT. We first introduce our experimental setup, then we show the effectiveness of our methodology by comparing it to state-of-the-art methods and ablation studies.

\subsection{Experimental Setup}

\paragraph{Datasets}
We conduct experiments on two widely used benchmark datasets for multi-task dense prediction, PASCAL-Context \cite{pc}, and NYUD-v2 \cite{nyud}. PASCAL-Context is based on the PASCAL VOC dataset \cite{voc}. It contains 4998 training and 5105 testing images and provides labels for five tasks of edge detection ('Edge'), semantic segmentation ('SemSeg'), human parts segmentation ('Parts'), surface normal estimation ('Normal'), and saliency detection ('Sal'). Among them, the surface normal and saliency labels are supplemented by \textit{Maninis} \textit{et al.} \cite{astmt} through label-distillation.
NYUD-v2 dataset consists of 795 training and 654 testing images of indoor scenes, with labels for four tasks of edge detection, semantic segmentation, surface normal estimation, and depth estimation ('Depth').
We follow the typical data augmentation pipeline in existing methods \cite{rcm, tsn}, including scaling, cropping, horizontal flipping and color jittering.

\begin{figure}
    \centering
    \includegraphics[width=0.9\linewidth]{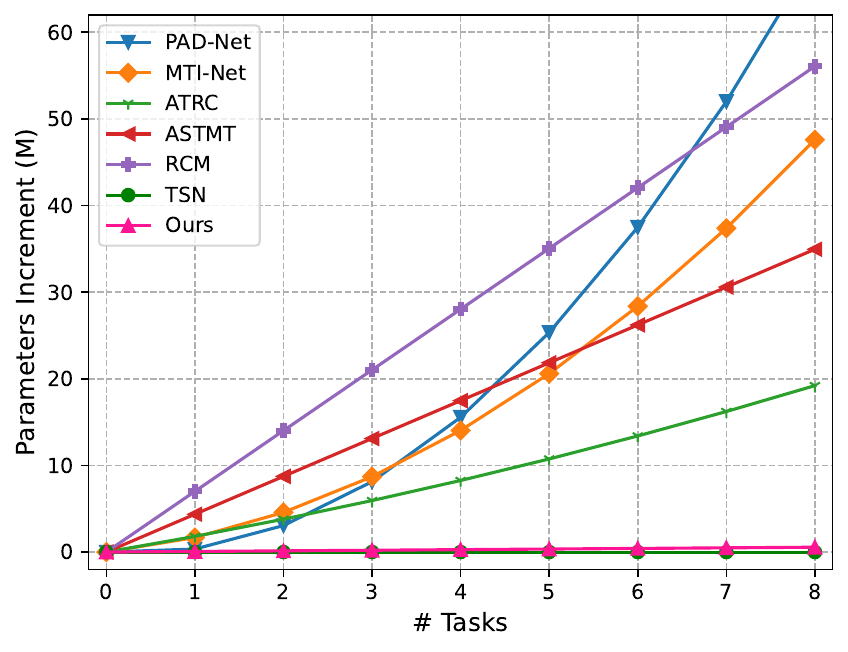}
    \caption{Increment of model parameters with additional tasks. It demonstrates the impact on the number of parameters added to a multi-task model as the number of tasks increases. Decoder-focused methods exhibit a quadratic relationship, and task-conditional methods demonstrate a linear relationship. Our model has a minor increment that shows high efficiency.}
    \label{fig:param}
\end{figure}

\paragraph{Implementation}
We test our method using Swin Transformer \cite{swin} backbones pre-trained on ImageNet-22K. We set prompt length $N_p$ to 10 in all PcT blocks and use Swin-T as the encoder backbone and four transformer blocks in the decoder unless otherwise stated.
The model is trained for 300 epochs on PASCAL-Context and 500 epochs on NYUD-v2, with a batch size of 8. We use AdamW optimizer \cite{adamw} with a learning rate of $10^{-4}$ and a weight decay rate of $10^{-4}$ and a cosine decay learning rate scheduler \cite{sgdr} with a warmup of 5 epochs. We implement our method with PyTorch \cite{pytorch} and run all experiments on 4 NVIDIA RTX3090 GPUs.

\paragraph{Baselines}
We construct strong single task baselines and multi-task baselines. They have the same architecture as our PGT except that they use original Swin Transformer blocks instead of PcT blocks. The single task baselines train each task with an individual network, while the multi-task baselines use the conventional multi-decoder structure.

\paragraph{Evaluation metric}
We follow the standard practice in evaluation metrics. We use optimal-dataset-scale F-measure (odsF) for edge detection, mean Intersection over Union (mIoU) for semantic segmentation, human parts segmentation and saliency detection, mean error (mErr) of predicted angles for surface normal estimation, and Root Mean Square Error (RMSE) for depth estimation.
To quantify the overall performance of a multi-task approach, we adopt the average per-task performance drop with respect to the single task baseline:
\begin{equation}
    \Delta_m=\frac{1}{N}\sum_{i=1}^N (-1)^{l_i}\frac{M_{m,i}-M_{b,i}}{M_{b,i}}
\end{equation}
where $N$ represents the number of tasks, $M_{m,i}$ and $M_{b,i}$ are the measure of task $i$ for the multi-task method and single task baseline, respectively. In addition, $l_i=1$ if a lower value denotes better for task $i$, and 0 otherwise.

\subsection{Efficiency Analysis}
Since model efficiency is a crucial consideration in our motivation, we first investigate on parameter efficiency. Fig. \ref{fig:param} illustrates how parameter sizes varies with the number of tasks. We can see that PAD-Net \cite{pad-net}, MTI-Net \cite{mti-net}, and ATRC \cite{atrc}, which are decoder-focused methods, exhibit quadratic curves. This is because these methods concentrate on extracting cross-task relationships by applying separate modules for each pair of tasks, leading to a quadratic growth with the number of tasks.
Meanwhile, task-conditional methods ASTMT \cite{astmt} and RCM \cite{rcm} show linear relationships. But it is important to note that these two methods exhibit a high increasing rate and even require more parameters than decoder-focused methods in some cases. The reason behind this is that they adopt inefficient conditioning strategies. The task-specific modules may contain more parameters than the shared stem if there are numerous tasks. 
In contrast, TSN \cite{tsn} applies the task-conditional paradigm on a single unified network and maintains constant parameters, regardless of the number of tasks.

\begin{table*}[t]
    \setlength{\tabcolsep}{12pt}
    \renewcommand{\arraystretch}{1.5}
    \centering
    \caption{Comparison to state-of-the-art models on PASCAL-Context dataset. 'Dec' in 'Params' shows the percentage (\%) of parameters used in the decoder over all parameters. '$\uparrow$' means higher is better and '$\downarrow$' means lower is better. '$\Delta_m\%$' denotes the average performance drop with respect to our single task baseline (higher is better).}
    \begin{tabular}{ccccccccccc}
    \hline
        \multirow{2}{*}{Model} & \multirow{2}{*}{Backbone} & Params / Dec & Edge & SemSeg & Parts & Normal & Sal & \multirow{2}{*}{$\Delta_m\%\uparrow$}\\
         & & (M / \%) & (odsF)$\uparrow$ & (mIoU)$\uparrow$ & (mIoU)$\uparrow$ & (mErr)$\downarrow$ & (mIoU)$\uparrow$ \\
    \hline
        single task baseline & Swin-T & 140.8 / 2.3 & 76.51 & 67.70 & 66.14 & 13.37 & 65.33  & 0.00 \\
        multi-task baseline & Swin-T & 30.8 / 10.6 & 74.68 & 65.78 & 60.68 & 13.95 & 64.44 &  -3.84 \\
    \hdashline
        ASTMT \cite{astmt} & ResNet-50 & 49.4 / 13.6 & 72.40 & 68.00 & 61.12 & 14.68 & 65.71 & -4.35 \\
        RCM \cite{rcm} & ResNet-18 & 46.1 / 58.9 & 71.30 & 65.70 & 58.12 & 13.70 & 66.38 & -4.55\\
        TSN \cite{tsn} & ResNet-34 & 28.4 / 25.0 & 71.80 & 67.60 & 58.00 & 16.10 & 64.30 &  -8.12\\
        TSN \cite{tsn} & Swin-T & 39.1 / 29.6 & 74.04 & 67.30 & 61.11 & 14.55 & 64.29 & -4.37\\
        PGT (Ours) & Swin-T & 28.5 / \textbf{2.7} & 73.93 & 67.58 & 62.58 & 13.95 & 65.59 & \textbf{-2.57}\\
    \hline
    \end{tabular}
    \label{tab:pc}
\end{table*}

Our model is not entirely independent of the number of tasks, but the increment in the number of parameters is minor that has a negligible effect on the overall model size. Since most modules are shared among tasks, the task-specific parameters only consist of prompts and prediction heads. For instance, for Swin-T as the encoder backbone with $C=96$ and a decoder with four PcT blocks, a task prompt with 10 tokens yields 48K additional parameters for each task, accounting for only 0.17\% of the total model parameters. Our decoder design is also lightweight, as the MLP modules and the transformer layer in the decoder have 0.64M parameters, and each prediction head has around 23.4K parameters.

Based on our analysis of model efficiency, we can confidently claim that our proposed PGT is a parameter-efficient method in that our framework is lightweight and the total number of parameters is only slightly affected when adding more tasks.

\subsection{Comparison to State-of-the-art}
\begin{table*}[t]
    \setlength{\tabcolsep}{14pt}
    \renewcommand{\arraystretch}{1.5}
    \centering
    \caption{Comparison to state-of-the-art models on NYUD-v2 dataset.}
    \begin{tabular}{cccccccccc}
    \hline
        \multirow{2}{*}{Model} & \multirow{2}{*}{Backbone} &Params / Dec & Edge & SemSeg & Normal & Depth & \multirow{2}{*}{$\Delta_m\%\uparrow$}\\
        & & (M / \%) & (odsF)$\uparrow$ & (mIoU)$\uparrow$ & (mErr)$\downarrow$ & (RMSE)$\downarrow$ \\
    \hline
        single task baseline & Swin-T & 112.7 / 2.3 & 77.45 & 42.44 & 19.96 & 0.5988  & 0.00 \\
        multi-task baseline & Swin-T & 30.1 / 8.6 & 77.50 & 40.69 & 20.19 & 0.6002 & -1.36 \\
    \hdashline
        ASTMT \cite{astmt} & ResNet-50 & 45.0 / 13.8 & 74.50	& 32.16 & 23.18 & 0.5700 & -9.84 \\
        RCM \cite{rcm} & ResNet-18 & 39.0 / 55.6 & 68.44 & 34.20 & 22.41 & 0.5700 & -9.63 \\
        TSN \cite{tsn} & ResNet-18 & 18.3 / 39.3 & 67.90 & 25.90 & 26.10 & 0.7270 & -25.87 \\
        TSN \cite{tsn} & Swin-T & 39.2 / 29.8 & 75.69 & 32.38 & 22.25 & 0.6874 & -13.06 \\
        PGT (Ours) & Swin-T & 28.4 / \textbf{2.6} & 77.05 & 41.61 & 20.06 & 0.5900 & \textbf{-0.38} \\
    \hline
    \end{tabular}
    \label{tab:ny}
\end{table*}

In this section, we compare performance of our method with existing task-conditional methods, namely ASTMT \cite{astmt}, RCM \cite{rcm}, and TSN \cite{tsn}, all of which share the same MTL paradigm as our method. To ensure a fair comparison, we select models with backbones that have similar or greater parameter counts than ours. Furthermore, we re-implement TSN using the Swin-T backbone and report its performance on two datasets.

Table \ref{tab:pc} provides the comparison with state-of-the-art methods on the PASCAL-Context dataset. Our method consistently outperforms existing approaches, showcasing the best average performance, with respect to our single task baseline. Specifically, our method achieves a 1.78\% and 1.98\% improvement over two larger models, ASTMT and RCM, while utilizing only 57.7\% and 61.8\% of their respective parameters.
In comparison to TSN with the ResNet-34 backbone, which has a similar parameter count to ours, our PGT achieves a significant 5.55\% improvement.
Even when considering models with the same Swin-T backbone, our method surpasses TSN by 1.8\% while employing substantially fewer parameters. 
It is worth mentioning that our lightweight decoder accounts for only 2.7\% of the total model parameters, which is considerably smaller than existing methods, which may allocate even more than half of the parameters to the decoders.
These results clearly prove the superior capability of task-specific feature adaptation attained by our PcT blocks under the guidance of prompts.

We also evaluate the proposed model on the NYUD-v2 dataset, as shown in Table \ref{tab:ny}. The effectiveness of our PGT is further validated, as it outperforms ASTMT and RCM by a significant gap of over 9 points.
Moreover, our method exhibits superior performance compared to TSN with the Swin-T backbone, with an improvement of 12.68\% while utilizing only 28.4M parameters, compared to TSN's 39.2M parameters.

We also find that our multi-task baseline readily outperforms previous CNN-based models, indicating that the transformer layer empowers the network with stronger feature extraction and generalization capabilities compared to the convolution layer. Notably, PGT achieves approximately 4$\times$ fewer parameters in decoder while maintaining around 1\% higher accuracy compared to the multi-task baseline.

In conclusion, our experimental results and intuitive comparisons over performance and the number of parameters in Fig. \ref{fig:comp} solidly establish the effectiveness of our method in terms of both parameter efficiency and multi-task performance. 

\begin{table}[t]
    \renewcommand{\arraystretch}{1.5}
    \centering
    \caption{Ablation on PcT block usage.}
    \begin{tabular*}{\hsize}{@{}@{\extracolsep{\fill}}ccccccc@{}}
    \hline
        \multirow{2}{*}{Usage} &  Edge & SemSeg & Normals & Depth & \multirow{2}{*}{$\Delta_m\%\uparrow$}\\
         & (odsF)$\uparrow$ & (mIoU)$\uparrow$ & (mErr)$\downarrow$ & (RMSE)$\downarrow$ \\
    \hline
        \makebox[0.06\textwidth][c]{encoder} & 76.98 & 41.61 & 20.00 & 0.5947 & -0.52 \\
        decoder & 76.73 & 38.62	& 20.12	& 0.6323 & -4.08 \\
        both & 77.05 & 41.61 & 20.06 & 0.5900 & \textbf{-0.38} \\
    \hline
    \end{tabular*}
    \label{tab:ab_ed}
\end{table}

\begin{figure}
    \scriptsize{~~~~~~Image~~~~~~~~~~~~~Edge~~~~~~~~SemSeg~~~~~~~~Parts~~~~~~~~~Normal~~~~~~~~~Sal}
    \vspace{-3mm}
    
    \subfloat{\includegraphics[width=0.2\linewidth, valign=c]{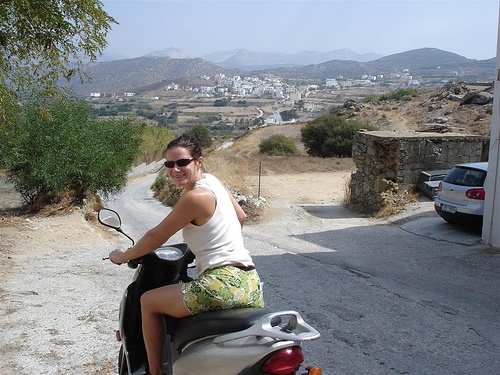}}
    \subfloat{\includegraphics[width=0.78\linewidth, valign=c]{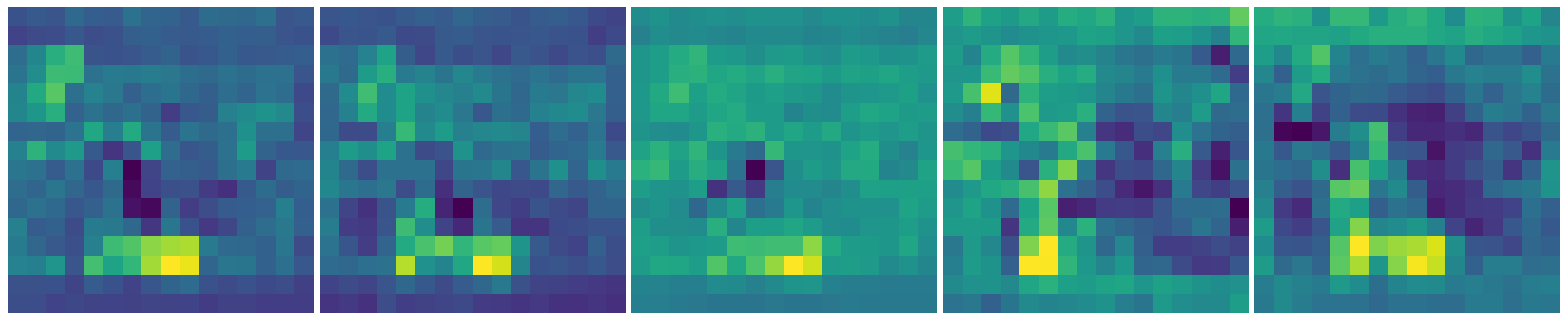}}
    \vspace{-4mm}

    \subfloat{\includegraphics[width=0.2\linewidth, valign=c]{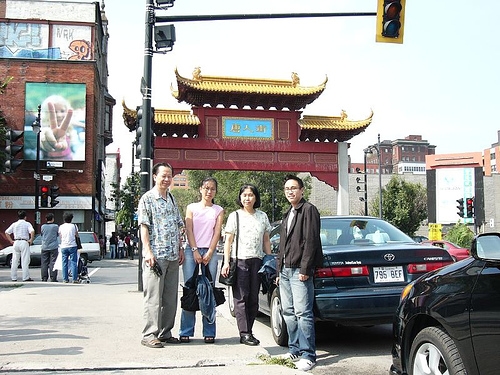}}
    \subfloat{\includegraphics[width=0.78\linewidth, valign=c]{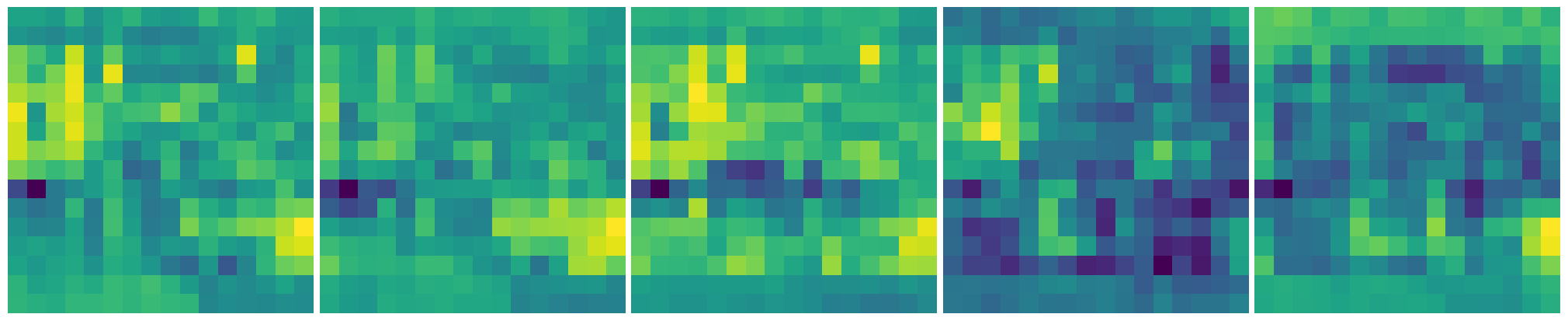}}
    \caption{Visualization of task-specific features extracted by task-conditional encoder. The heatmaps show different concentrations for different tasks.}
    \label{fig:vis_en}
\end{figure}

\begin{table}[t]
    \renewcommand{\arraystretch}{1.5}
    \centering
    \caption{Ablation on prompt length.}
    \label{tab:ab_promptlen}
    \begin{tabular*}{\hsize}{@{}@{\extracolsep{\fill}}ccccccc@{}}
    \hline
        \multirow{2}{*}{\# Prompts} &  Edge & SemSeg & Normal & Depth & \multirow{2}{*}{$\Delta_m\%\uparrow$}\\
         & (odsF)$\uparrow$ & (mIoU)$\uparrow$ & (mErr)$\downarrow$ & (RMSE)$\downarrow$ \\
    \hline
        \makebox[0.06\textwidth][c]{1} & 77.08 & 41.30 & 20.03 & 0.6091 & -1.31 \\
        2 & 77.02 & 41.90 & 19.93 & 0.6064 & -0.74 \\
        4 & 77.06 & 41.11 & 20.05 & 0.5968 & -0.94 \\
        10 & 77.05 & 41.61 & 20.06 & 0.5900 & \textbf{-0.38} \\
        20 & 76.77 & 41.14 & 20.02 & 0.6013 & -1.16 \\
    \hline
    \end{tabular*}
\end{table}

\begin{table}[t]
    \renewcommand{\arraystretch}{1.5}
    \centering
    \caption{Ablation on depths of decoder.}
    \begin{tabular*}{\hsize}{@{}@{\extracolsep{\fill}}ccccccc@{}}
    \hline
        \multirow{2}{*}{Depths} &  Edge & SemSeg & Normal & Depth & \multirow{2}{*}{$\Delta_m\%\uparrow$}\\
         & (odsF)$\uparrow$ & (mIoU)$\uparrow$ & (mErr)$\downarrow$ & (RMSE)$\downarrow$ \\
    \hline
        \makebox[0.06\textwidth][c]{2} & 76.93 & 41.17 & 19.88 & 0.5961 & -0.70 \\
        4 & 77.05 & 41.61 & 20.06 & 0.5900 & \textbf{-0.38} \\
        8 & 77.12 & 41.48 & 19.94 & 0.5961 & -0.53 \\
    \hline
    \end{tabular*}
    \label{tab:ab_depth}
\end{table}

\subsection{Ablation Study}
We conduct ablation studies to validate the proposed methodology on the NYUD-v2 dataset using a Swin-T backbone.

\paragraph{Effect of PcT blocks}
In Table \ref{tab:ab_ed}, we investigate the impact of using task-conditional modules, our PcT blocks, in the encoder and decoder separately to showcase the effectiveness of our proposed method. The results reveal that using PcT blocks solely in the decoder leads to a significant performance degradation, while using them exclusively in the encoder still yields competitive outcomes. As incorporating a task-conditional encoder has a more substantial effect on performance, it suggests that the PcT block provides the encoder with a solid capability to learn task-specific features and general representations, and enables the use of a lightweight decoder. Additionally, the task-conditional decoder proves beneficial in achieving higher accuracy across almost all tasks. 
To provide visual evidence, we depict two samples of task-specific features extracted by the task-conditional encoder in Fig. \ref{fig:vis_en}. We can observe a considerable variation in the concentration of different tasks, which supports the argument that diverse tasks have distinct requirements for feature extraction in the encoder.

These findings underscore the importance of incorporating task-conditional modules in both the encoder and decoder to fully leverage the advantages of our proposed PcT block. The feature adaptation guided by prompts significantly contributes to superior performance on various tasks, as demonstrated in the previous sections.

\paragraph{Prompt length}
In Table \ref{tab:ab_promptlen}, we study the influence of different choices for the prompt length, which is proportional to the number of task-specific parameters. We find that the best performance is achieved when the prompt length is set to 10, and adding more prompts does not provide any additional benefit. Potentially, models with an excessive number of prompts may overly prioritize task-specific representations, at the expense of weakening the acquisition of universal information shared across all tasks.
Notably, our method outperforms existing task-conditional methods even with a prompt length of 1. This surprising result indicates that our conditioning strategy through prompting is superior to previous approaches, such as residual adapter \cite{astmt} and convolution reparameterization \cite{rcm}, though our method introduces only a tiny number of task-specific parameters. Based on these results, we choose a prompt length of 10 as default setting.

\begin{table}[t]
    \setlength{\tabcolsep}{4pt}
    \renewcommand{\arraystretch}{1.5}
    \centering
    \caption{Ablation on backbones of encoder.}
    \begin{tabular*}{\hsize}{@{}@{\extracolsep{\fill}}ccccccc@{}}
    \hline
        \multirow{2}{*}{Backbone} & \multirow{2}{*}{Model} &  Edge & SemSeg & Normals & Depth & \multirow{2}{*}{$\Delta_m\%\uparrow$}\\
         & & (odsF)$\uparrow$ & (mIoU)$\uparrow$ & (mErr)$\downarrow$ & (RMSE)$\downarrow$ \\
    \hline
        \multirow{3}{*}{Swin-S} & single task & 78.17 & 47.89 & 19.47 & 0.5533 & 0.00 \\
        & multi-task & 78.28 & 45.27 & 19.54 & 0.5790 & -2.58 \\
        & Ours & 78.04 & 46.43 & 19.24 & 0.5468 & \textbf{-0.21} \\
    \hdashline
        \multirow{3}{*}{Swin-B} & single task & 78.37 & 46.99 & 19.38 & 0.5559 & 0.00 \\
        & multi-task & 78.75 & 46.05 & 19.43 & 0.5737 & -1.24 \\
        & Ours & 78.28 & 47.42 & 19.12 & 0.5502 & \textbf{0.79} \\
    \hline
    \end{tabular*}
    \label{tab:ab_backbone}
\end{table}

\begin{figure}
    \centering
    \subfloat[First layer in encoder]{\includegraphics[width=0.5\linewidth]{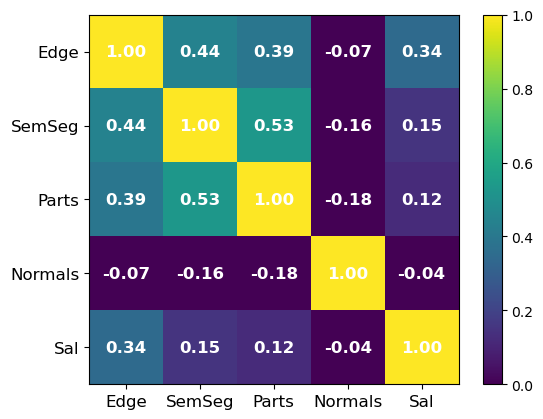}}
    \subfloat[Layer in decoder]{\includegraphics[width=0.5\linewidth]{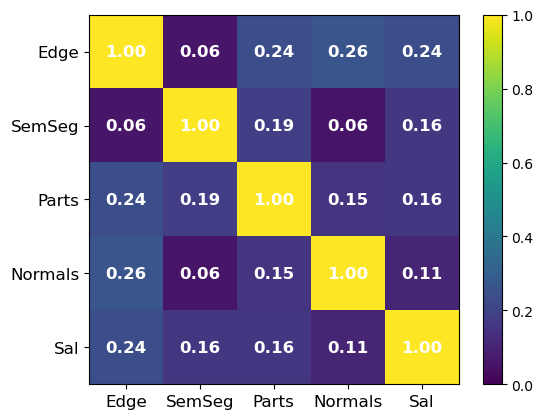}}
    \caption{Similarity of learned prompts from different layers in model trained on PASCAL-Context dataset.}
    \label{fig:prompt_rela}
\end{figure}

\paragraph{Depths of the decoder}
Our decoder uses four PcT blocks in default to implement an effective yet lightweight decoder. In Table. \ref{tab:ab_depth}, we ablate on the depth of the decoder to see how the number of blocks used influences the overall performance. We observe a 0.32\% gain when the depth is increased from two to four. 
Nevertheless, as the depth continues to increase, the accuracy does not improve and even decreases in semantic segmentation and depth estimation, while the parameters and computational costs keep rising. Hence, we choose to set the depth of the transformer layer in the decoder to four, resulting in a more efficient model.

\paragraph{Effect of different encoder backbones}
We compare the performance of our model when employing different encoder backbones, and the results are listed in Table. \ref{tab:ab_backbone}. We can observe that the absolute performance consistently improves on most tasks when using backbones with larger capacity, especially on semantic segmentation accuracy.
Moreover, our model performs closely to the single task baseline with Swin-S and exceeds the Swin-B baseline, which has nearly five times more parameters than ours. Our method also outperforms the multi-task baselines by a clear margin in nearly all tasks, demonstrating the effectiveness and robustness of our approach in generalization.

\paragraph{Prompts relationships}
In order to gain deeper insights into task prompts, we calculate the cosine similarity of learned prompts for each task pair, which allows us to determine the relationships between different tasks. For the PASCAL-Context dataset, we visualize the prompt relationships derived from the first transformer layer in the encoder and the transformer layer in the decoder in Fig. \ref{fig:prompt_rela}. 
We can observe that the learned prompts in the first encoder layer exhibit task relationships that align with our intuitive understanding. For instance, semantic segmentation and human parts segmentation, both being segmentation tasks, demonstrate a close association. 
Additionally, they are also closely related to edge detection since edges often coincide with the boundaries of segmentation masks. Moreover, there exists a noticeable distinction between surface normal estimation and other tasks. This discrepancy can be attributed to the fact that surface normal estimation is a 3D task, while the others are 2D tasks. 
Conversely, in the decoder layer, we observe a decrease in the interrelatedness of the task prompts.
This observation suggests that the prompts become more task-specific, guiding the transformer layer to generate unique features that fulfill the requirements of each specific task.  

\subsection{Qualitative results}
\begin{figure}[t]
    \scriptsize{~~~~~~~~Image~~~~~~~~~~Edge~~~~~~~~~SemSeg~~~~~~~~~Parts~~~~~~~~~Normal~~~~~~~~~~Sal}
    \vspace{-3mm}
    
    \subfloat{
        \rotatebox{90}{\scriptsize{~~~~Ground Truth}}
        \includegraphics[width=0.15\linewidth]{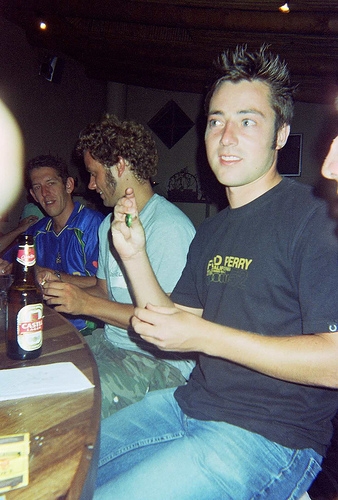}
        \includegraphics[width=0.15\linewidth]{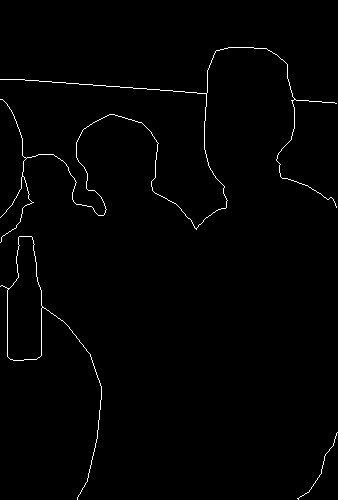}
        \includegraphics[width=0.15\linewidth]{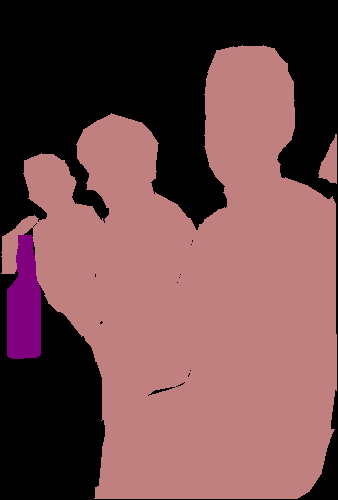}
        \includegraphics[width=0.15\linewidth]{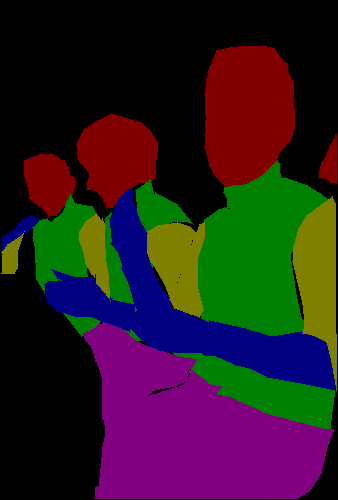}
        \includegraphics[width=0.15\linewidth]{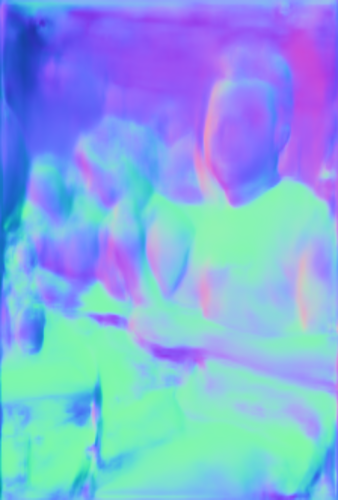}
        \includegraphics[width=0.15\linewidth]{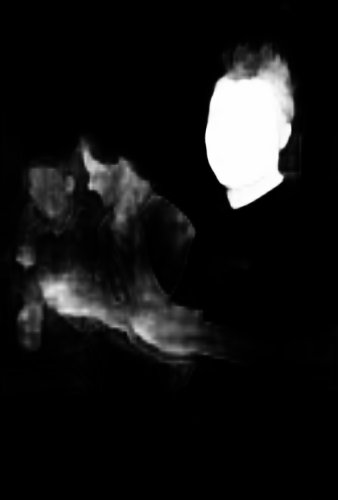}
    }

    \vspace{-3mm}

    \subfloat{
        \rotatebox{90}{\scriptsize{~~~~~~~~RCM}}
        \includegraphics[width=0.15\linewidth]{2008_000034.jpg}
        \includegraphics[width=0.15\linewidth]{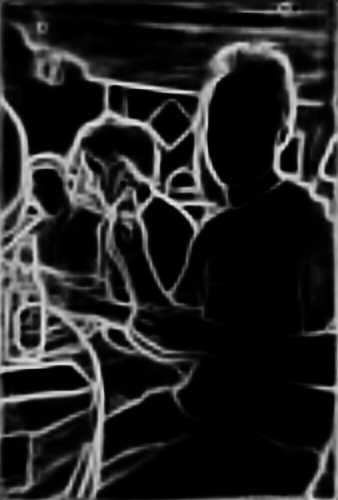}
        \includegraphics[width=0.15\linewidth]{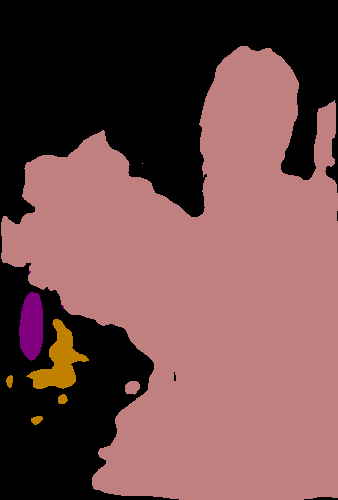}
        \includegraphics[width=0.15\linewidth]{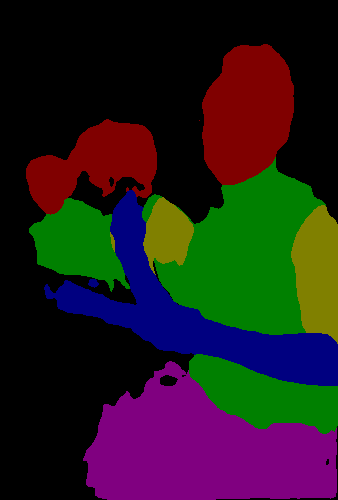}
        \includegraphics[width=0.15\linewidth]{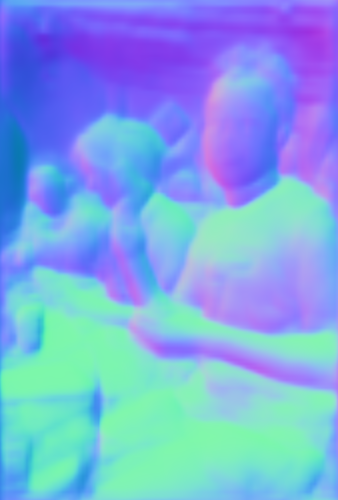}
        \includegraphics[width=0.15\linewidth]{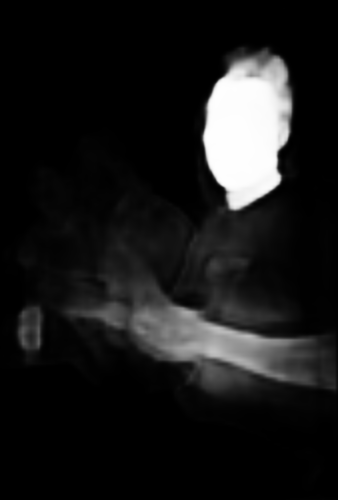}
    }

    \vspace{-3mm}

    \subfloat{
        \rotatebox{90}{\scriptsize{~~~~~~~~TSN}}
        \includegraphics[width=0.15\linewidth]{2008_000034.jpg}
        \includegraphics[width=0.15\linewidth]{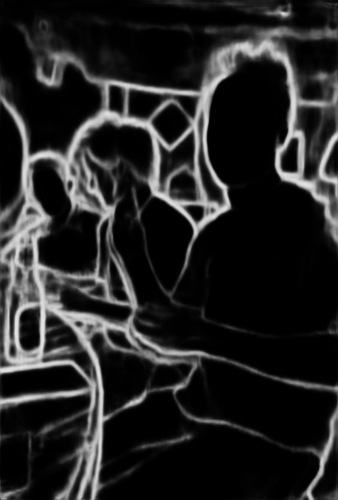}
        \includegraphics[width=0.15\linewidth]{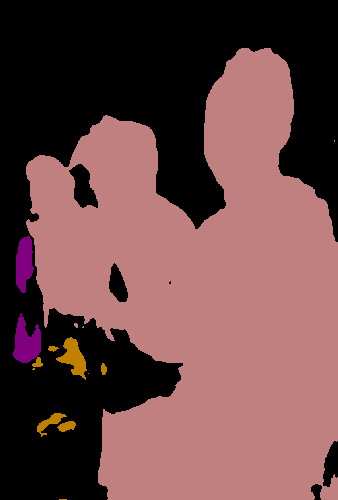}
        \includegraphics[width=0.15\linewidth]{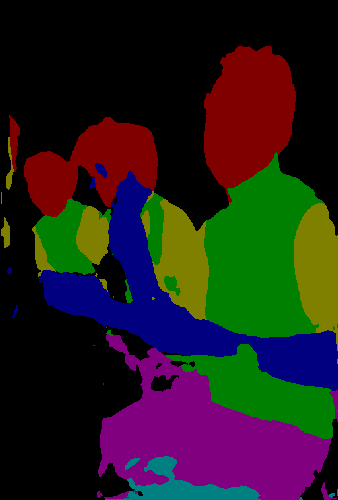}
        \includegraphics[width=0.15\linewidth]{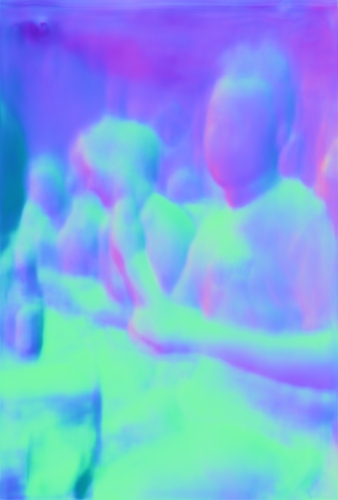}
        \includegraphics[width=0.15\linewidth]{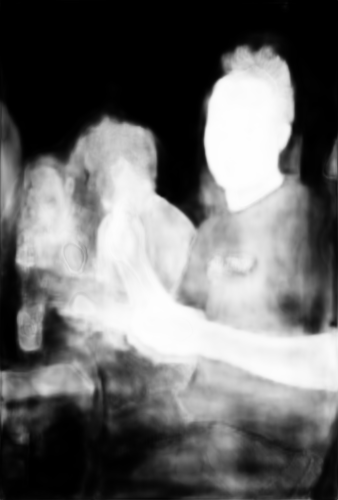}
    }

    \vspace{-3mm}

    \subfloat{
        \rotatebox{90}{\scriptsize{~~~~~~~~Ours}}
        \includegraphics[width=0.15\linewidth]{2008_000034.jpg}
        \includegraphics[width=0.15\linewidth]{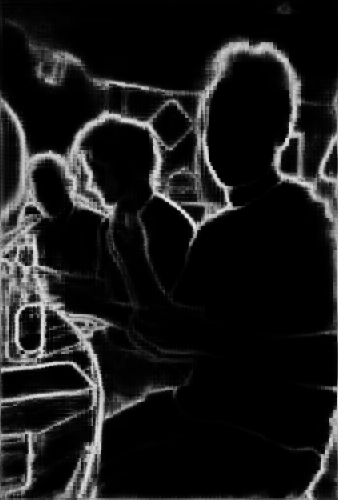}
        \includegraphics[width=0.15\linewidth]{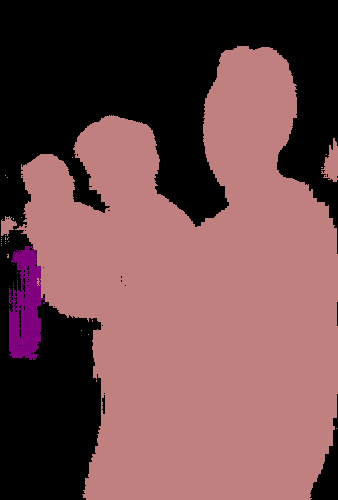}
        \includegraphics[width=0.15\linewidth]{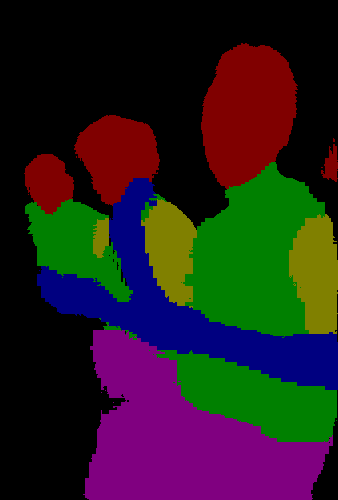}
        \includegraphics[width=0.15\linewidth]{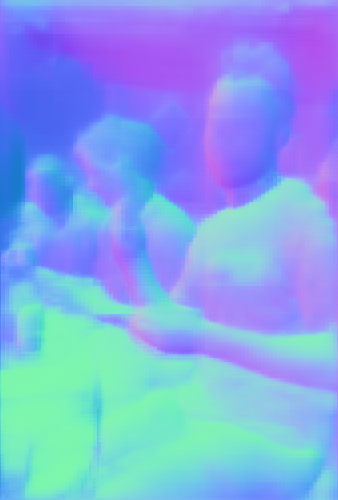}
        \includegraphics[width=0.15\linewidth]{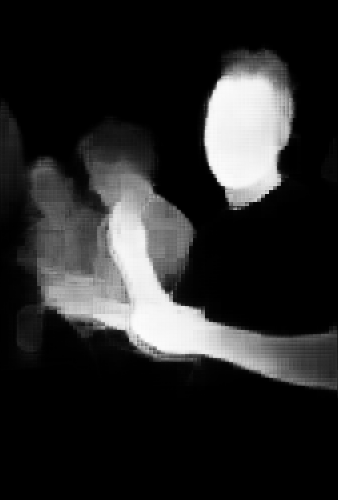}
    }
    \caption{Qualitative results compared with RCM and TSN on PASCAL-Context dataset.}
    \label{fig:qual}
\end{figure}

To intuitively compare the proposed PGT with existing task-conditional methods, we visualize the task predictions of RCM, TSN, and our model with an example from the PASCAL-Context dataset in Fig. \ref{fig:qual}. Our model obviously generates better details with less error, especially in semantic segmentation and human parts segmentation results, and more closely resembles the ground truths in other tasks.

\section{Conclusion}
In this paper, we introduce Prompt Guided Transformer (PGT), a novel method for multi-task dense prediction. PGT adopts a task-conditional framework and leverages a transformer-based model to improve multi-task performance while addressing the parameter challenge. We propose the integration of Prompt-conditioned Transformer (PcT) blocks in both the encoder and decoder, promoting representation learning and parameter-efficient adaptation across tasks under the guidance by task-specific prompts. Through extensive experiments conducted on the PASCAL-Context and NYUD-v2 datasets, we demonstrate that PGT achieves significant performance improvements over previous task-conditional models while using fewer parameters, thus validating the effectiveness of our conditioning strategy and proposed model.
Moreover, our method allocates the smallest proportion of parameters to the decoder, further underscoring its efficiency.   
In future work, we plan to explore approaches to reduce computational costs in task-conditioning and further enhance overall performance.

\section*{Acknowledgment}
This paper is supported by National Nature Science Foundation of China (62176155, 62066002), Shanghai Municipal Science and Technology Major Project (2021SHZDZX0102).

\bibliographystyle{IEEEtranS}
\bibliography{egbib}

\begin{IEEEbiography}[{\includegraphics[width=1in,height=1.25in,clip,keepaspectratio]{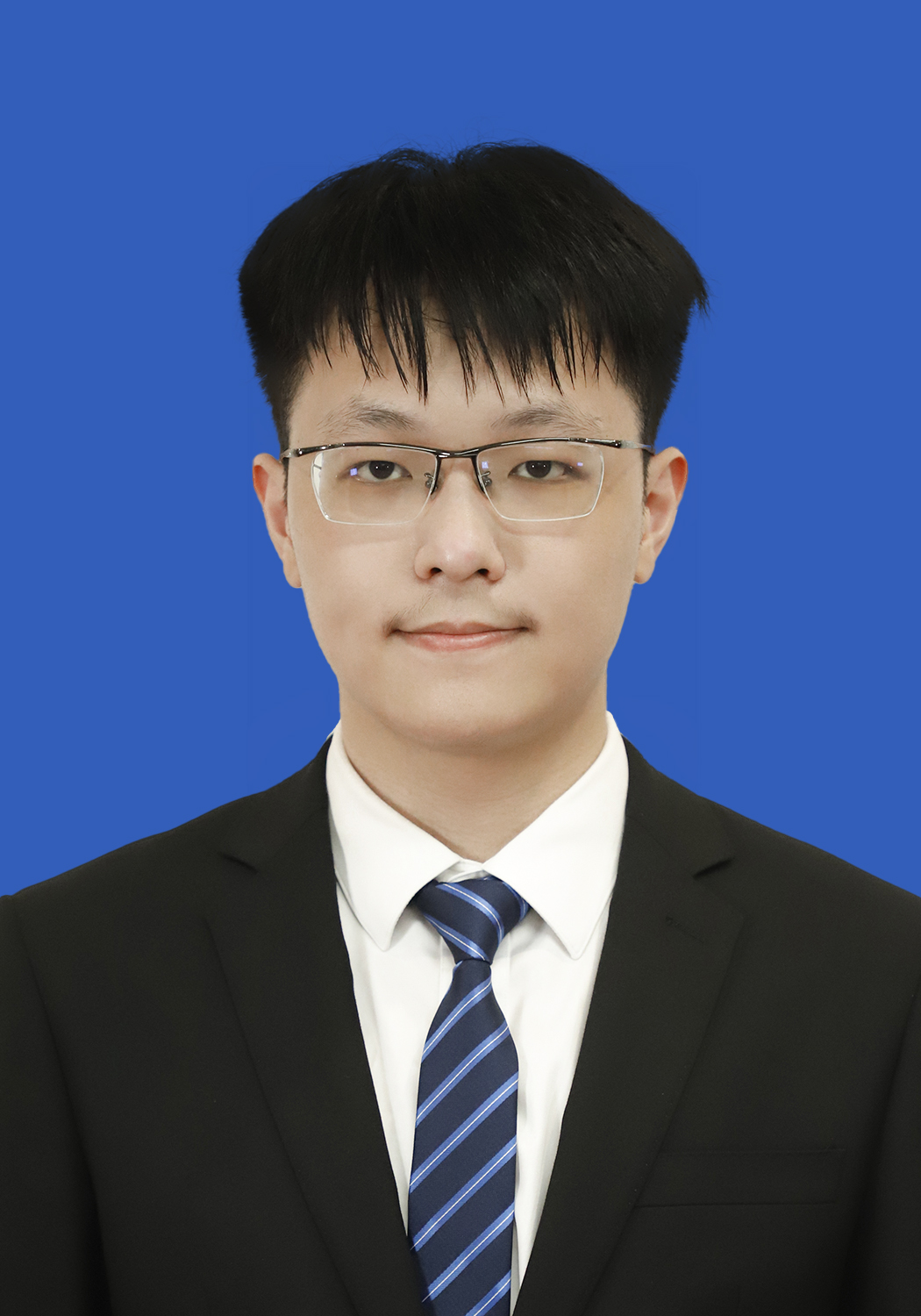}}]{Yuxiang Lu}
is a Master student at the Department of Computer Science and Engineering, Shanghai Jiao Tong University. He received the B.Eng. from Shanghai Jiao Tong University. His research interests includes computer vision, multi-task learning, and deep learning.
\end{IEEEbiography}

\begin{IEEEbiography}[{\includegraphics[width=1in,height=1.25in,clip,keepaspectratio]{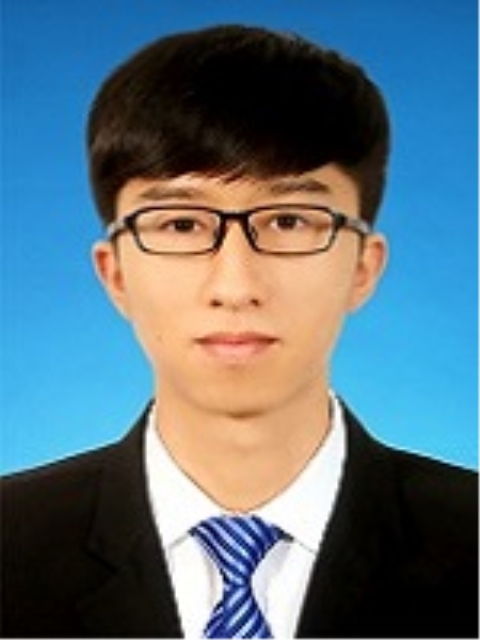}}]{Salayiding Sirejiding}
is a Ph.D student at the Department of Computer Science and Engineering, Shanghai Jiao Tong University. He received the B.Eng. from Harbin Institute of Technology, received the M.S. degree at Shanghai Jiao Tong University. His research interests include computer vision, multi-task learning and machine learning.
\end{IEEEbiography}

\begin{IEEEbiography}[{\includegraphics[width=1in,height=1.25in,clip,keepaspectratio]{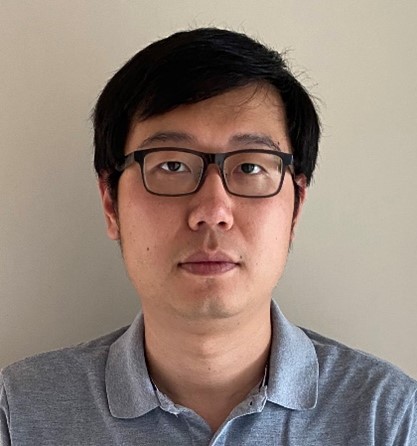}}]{Yue Ding}
received the Ph.D. from Shanghai Jiao Tong University in 2018. He is now an assistant research professor at the Department of Computer Science and Engineering, Shanghai Jiao Tong University. His main research interests are recommender systems, graph neural networks, sequence modeling, and data mining.
\end{IEEEbiography}

\begin{IEEEbiography}[{\includegraphics[width=1in,height=1.25in,clip,keepaspectratio]{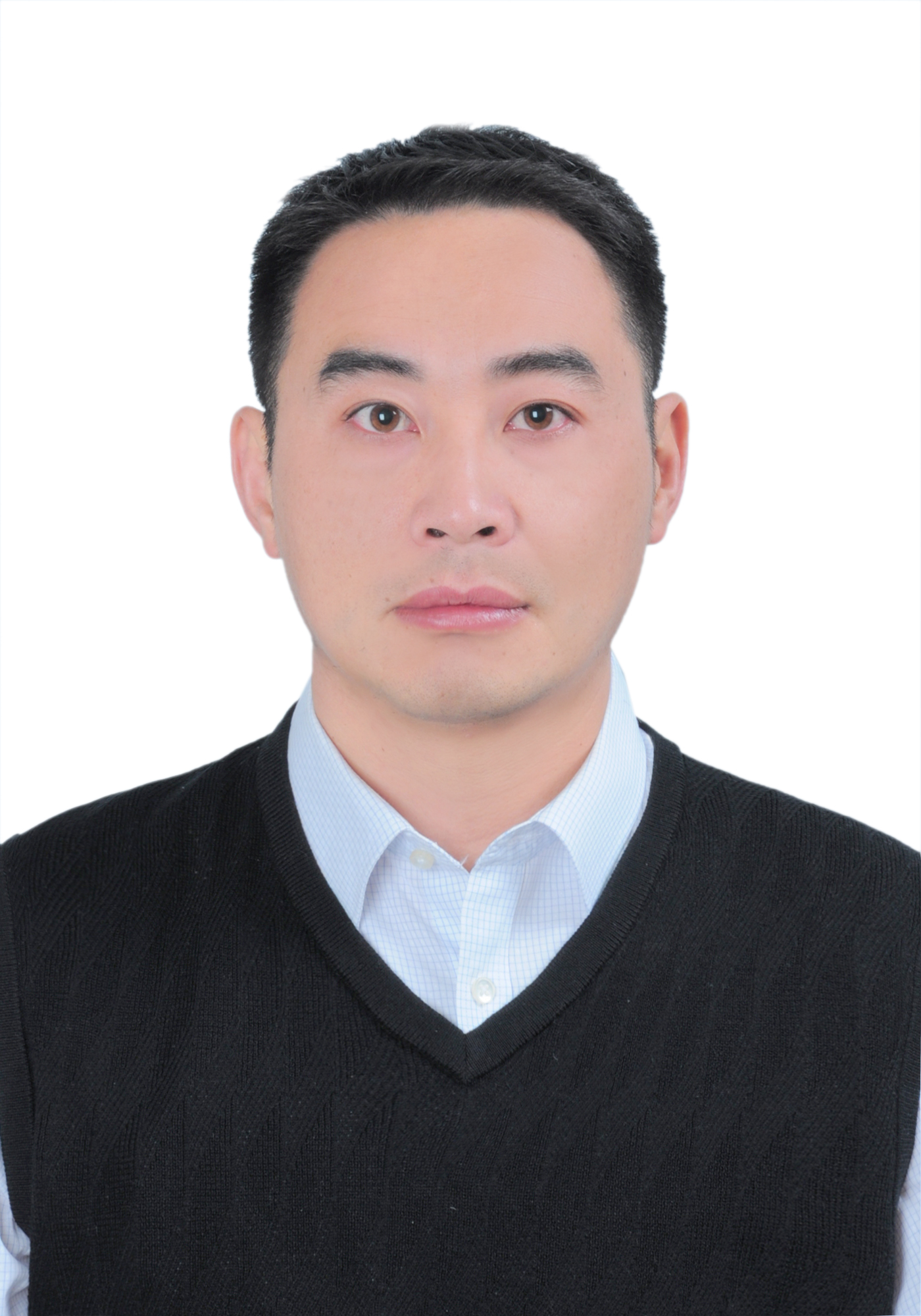}}]{Chunlin Wang}
is a professor of the  School of Information Science and Technology, Chuxiong Normal University, Chuxiong, China. His research interests include machine learning,computer vision. He has authored or co-authored more than 10 research papers in journals and premier conferences. He has received 3 projects funded by the NSFC and the Yunnan Provincial Government.
\end{IEEEbiography}

\begin{IEEEbiography}[{\includegraphics[width=1in,height=1.25in,clip,keepaspectratio]{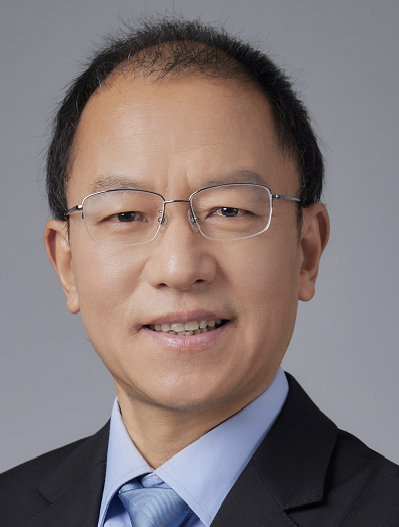}}]{Hongtao Lu}
Hongtao Lu is a Professor with the Department of Computer Science and Engineering, Shanghai Jiao Tong University, Shanghai, China. 
His research interests include machine learning, deep learning, computer vision, and pattern recognition. 
He has authored or co-authored more than 100 research papers in journals and premier conferences such as IEEE TRANSACTION, Pattern Recognition, CVPR, ECCV, AAAI etc. 
His authored or co-authored papers have gotten more than 1300 citations from Web of Science, and more than 7000 citations from Google Scholar, his H-index is 48. 
He was PIs for dozens of projects from NSFC, Ministry of Science and Technology, Ministry of Education.Municipal Government of Shanghai, and industries. He has been continuously listed among the most cited Researchers in computer science in China by Elsevier from years of 2014 to 2018. He also got several research awards from the government.
\end{IEEEbiography}

\end{document}